\documentclass[hidelinks]{scrartcl}
\usepackage[backend=bibtex,natbib=true,url=false,style=authoryear-ibid]{biblatex}
\usepackage{mathrsfs}
\usepackage{graphicx}
\usepackage[font=scriptsize,labelfont=bf]{caption}
\usepackage{hyperref}
\hypersetup{
	colorlinks = true,
    linkcolor = blue,
    urlcolor  = blue,
    citecolor = blue,
    anchorcolor = blue]
}
\usepackage{paralist}
\usepackage[textsize=footnotesize]{todonotes}

\bibliography{../biblio/weaver, ../biblio/vveitas}

\begin{document}
\title{Open Ended Intelligence}
\subtitle{\LARGE The individuation of Intelligent Agents}
\author{\large David Weinbaum (Weaver) (\texttt{space9weaver@gmail.com})\\
\large Viktoras Veitas (\texttt{vveitas@gmail.com})\\
\large The Global Brain Institute, VUB}
\date{\large May, 2015}

\maketitle
\begin{abstract}
\footnotesize{Artificial General Intelligence (AGI) is a field of research aiming to distill the principles of intelligence that operate independently of a specific problem domain or a predefined context and utilize these principles in order to synthesize systems capable of performing any intellectual task a human being is capable of and eventually go beyond that. While ``narrow'' artificial intelligence that focuses on solving specific problems such as speech recognition, text comprehension, visual pattern recognition, robotic motion, etc. has shown quite a few impressive breakthroughs lately, understanding general intelligence remains elusive. 

In this paper we offer a novel theoretical approach to understanding general intelligence. We start with a brief introduction of the current conceptual approach. Our critique exposes a number of serious limitations that are traced back to the ontological roots of the concept of intelligence. We then propose a paradigm shift from intelligence perceived as a competence of individual agents defined in relation to an \textit{a priori} given problem domain or a goal, to intelligence perceived as a formative process of self-organization by which intelligent agents are individuated. We call this process \textit{open-ended intelligence}. This paradigmatic shift significantly extends the concept of intelligence beyond its current conventional definitions and overcomes the difficulties exposed in the critique.  

Open-ended intelligence is developed as an abstraction of the process of cognitive development so its application can be extended to general agents and systems. We introduce and discuss three facets of the idea: the philosophical concept of individuation, sense-making -- the bringing forth of a world of objects and relations, and the individuation of general cognitive agents in the light of the enactive approach to cognition and assemblage theory. We study these in order to establish in what sense formative individuating processes are indeed intelligent and why they are open-ended.
 
We further show how open-ended intelligence can be framed in terms of a distributed, self-organizing network of interacting elements (i.e. a complex adaptive system) and how such a process is scalable. The framework highlights an important relation between coordination and intelligence and a new understanding of values. We conclude with a number of questions for future research.\\

\raggedleft{\textbf{Keywords:} intelligence, cognition, individuation, assemblage, self-organization, sense-making, coordination, enaction}}

\end{abstract}

\section{Introduction -- Intelligence and networks}
We live in the age of networks: ecological networks, biological networks, digital networks, logistic networks, knowledge networks, social networks and so on. It is an age of plurality, of diversity and above all, of interconnectedness. The internet, the most prominent actual exemplar of these concepts, is not only transforming the way we live and interact in the everyday, but furthermore has engendered a powerful image in our minds -- the image of the network. This image has already a strong grasp over both the way we reason and our imagination. In that, it sets the horizons of possible invention \citep{hui_collective_2013}. Deploying networks as an explanatory platform for cognition and intelligent behavior is an established practice in computational neuroscience \citep{edelman_universe_2000, tononi_consciousness_2008}, general cognitive science \citep{bechtel_connectionism_2002} and other fields. The relations between the network concept and  intelligence are many and strong. Primary of which is the fact that brains, the most advanced intelligent machines we know about as of today, are vast networks of interconnected neurons. The field of ``narrow'' artificial intelligence (AI) that focuses on goal-specific kinds of intelligence such as speech recognition, text comprehension, visual pattern recognition, robotic motion, etc. has known quite a few impressive breakthroughs lately. The highly competent AI agents developed today rely heavily on vast networks of artificial neurons. Their construction is inspired by biological brains and their competences begin to rival those of humans in addressing specific problems.   

The field of Artificial General Intelligence (AGI) is much more ambitious in comparison. It aims to distill the principles of intelligence that operate independently of a specific problem domain or a predefined context and utilize these principles to synthesize machines capable of performing any intellectual task a human being is capable of and eventually go beyond that. There is no doubt that the network concept holds powerful keys to understanding general intelligence and to the vision of building AGI agents. The goal of this paper is to examine, from a philosophical perspective, the conceptual foundations of intelligence and their emergence in the dynamics of distributed, disparate, interconnected structures\footnote{The allusion to the internet has already captured the popular imagination. Many believe that one day in the foreseeable future the internet, will `awaken' and become a conscious aware super-intelligent entity. Some even claim that this is already happening.}.

The following section briefly introduces the current conceptual approach to General Intelligence and criticizes it. We expose a number of implicit hidden assumptions that the definition of AGI is based upon. These assumptions place \textit{a priori} conceptual limits on how `general' general intelligence can be. Section \ref{sec:ontological} is a philosophical exploration of the ontological roots of intelligence and presents the theory of individuation, providing an alternative concept of intelligence as a process and not as a given competence. This novel approach overcomes the difficulties exposed in section \ref{sec:problems} and significantly extends the concept beyond the definition in \ref{subsec:current_definition}. As the title of the paper suggests, the term \textit{open-ended intelligence} will be used to describe intelligence as a process. In a nutshell, intelligence is the process of bringing forth a world of objects and their relations, or in other words, a continuous process of sense-making. Section \ref{sec:individuation_of_cognition} discusses how the  process of individuation is applied to cognition as an ongoing sense-making activity. An important theoretical bridge is made between the concept of individuation and sense-making as an actual process of \textit{cognitive development}. It is in the cognitive development of systems that open-ended intelligence is manifested. Here, by constructing a descriptive framework of the individuation of cognition, we study the various facets and implications of applying our approach and in what sense the formative individuating processes discussed are considered intelligent. We conclude with a list of open questions and issues for further research.   

\section{Conceptual problems with General Intelligence} \label{sec:problems}
\subsection{Definition of General Intelligence}\label{subsec:current_definition}
Intelligence is a difficult concept to define, especially in its general, context-independent sense. Many different context-bound definitions do exist however in diverse disciplines such as psychology, philosophy of mind, engineering, computer science, cognitive science and more. It is far from simple, if at all possible, to reach a common-ground definition that transcends the epistemological barriers between disciplines. \citet{legg_collection_2007} have compiled the most comprehensive collection to date of definitions of intelligence. A shorter review of various representative examples of such definitions can be found in \citep{legg_machine_2008}. Based on this broad review and their attempt to found a formal theoretical approach to general intelligence, Hutter and Legg have distilled the following definition: 
\begin{quotation} 
	Intelligence measures an agent's ability to achieve goals in a wide range of environments. \citep[p. 6]{legg_machine_2008}
\end{quotation} 
This definition tries to capture the broadest possible consideration of goals and operating environments. \citet{goertzel_cogprime:_2012} uses a slightly different version emphasizing the pragmatic real-world ``ability to achieve complex goals in complex environments'', something that is somewhat lost in Hutter's AIXI all-encompassing design \citep{hutter_universal_2005}. Still, from a foundational point of view, these two versions are in agreement. It would therefore be a good starting point to expose the problematic nature of such definition.

\subsection{Criticism of the definition} \label{subsec: criticism}
\begin{figure}[h]
	\centering
 	\includegraphics[scale=0.4]{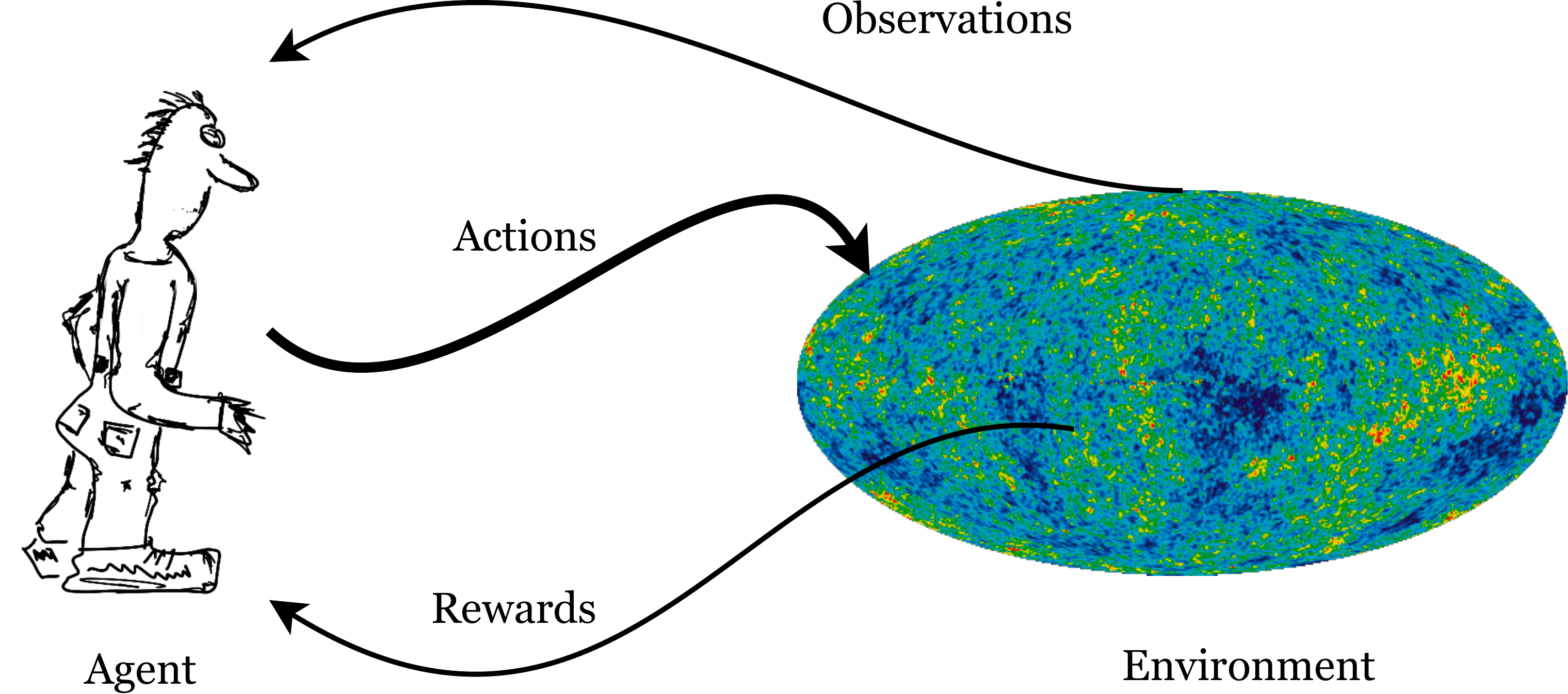}   	
  	\caption{Agent - Environment relations}
  	\label{fig:Agent-Environment}
\end{figure}

Figure \ref{fig:Agent-Environment} depicts a scheme of the agent--environment model that is the basis for the above definition. The story that goes with the scheme is that the agent, based on a flow of observations it receives from the environment, engages in a flow of actions made to achieve an optimized flow of rewards. The intelligence of the agent is a measure of its competence to match actions to observations such that it will achieve high rewards in a variety of diverse environments.         
With this definition a few presumptions are already clearly apparent:
\begin{description}
	\item[The~agent~environment distinction]-- The first strong assumption is that the agent is clearly distinct from its environment. It has a well defined contour across which it interacts with the environment. Additionally, the contour implicitly defines the kinds of interaction that can take place between the agent and the environment.
	\item[The~environment]-- The status of the environment is problematic in two aspects: first, due to the hidden assumption about the \textit{a priori} givenness of the environment and second, due to the assumption about its observer independent status. In Hutter's AIXI model \citep{hutter_universal_2005}, Solomonoff-Levin universal prior distribution \citep{solomonoff_formal_1964, solomonoff_formal_1964-1,legg_machine_2008} is a minimal knowledge predictor of the environment's behavior in the  most general case. It incorporates both Epicurus' principle and the principle of Occam's razor \citep[chap 2.]{legg_machine_2008} and describes the agent's best guess given its initial ignorance regarding the environment. Using the universal prior as a basis, the agent can reliably induce the future distribution of behaviors of the environment as more data on its behavior becomes available. But the subject matter of universal induction is only the agent's knowledge of the environment. The universal prior and the method of Bayesian induction assume an \textit{a priori} given environment with an observer independent status. Induction, therefore, only means the effective reduction of the observer's ignorance regarding the environment. Moreover, the actions of the agent can only affect the environment within its already given definitional constraints. The agent cannot \textit{change} the environment -- only discover its behavior and respond. Agent-environment reflexivity, which is so apparent in actual systems, is either highly ambiguous or entirely left out.      
	\item[Goal~driven~reward]-- Clearly the environment does not `give' rewards, as it is sometimes implied by considering it as an agent, only that certain states of the environment are more favorable than others relative to the agent's goals and in the context of its current internal state. For the definition of intelligence to be operative, it must therefore involve yet another presumption, namely, that the agent possesses a clearly defined goal (or a set of goals) that maps values to both internal and environment states. In its actions, the agent attempts to move the environment from its current given state to the state that is most favorable in terms of rewards given the dynamic context of its internal state. 
	\item[The~agent's~capacities]-- It is further implied that the agent is somehow structured by past interactions with the environment (knowledge) and has a computing capacity that affords the matching of actions to observations and the evaluation of rewards relative to its goals in the context of its state. This is of course a robust common sense assumption but as argued about the environment, presuming an agent endowed with \textit{a priori} given general capacities, leaves a lot out of the equation. After all, intelligent agents do not spontaneously appear ready-made in some purely conceptual space. Some evolutionary process is necessarily involved and must not be overlooked.    	   
\end{description}
All the presumptions listed here appeal strongly to common sense and frame the concept of intelligence in a reasonable and pragmatic manner. However, they also limit the generality of the concept in a few profound ways: 
\begin{itemize}
	\item  Processes of differentiation and boundary formation that determine the agent--environment distinctions are excluded. Such processes that can be broadly categorized as processes of self-organization can be gradual and possibly express intelligence of a kind that is not considered by the definition in \ref{subsec:current_definition}.
	\item  	Processes that are not clearly defined \textit{a priori} in terms of their goals and derived values are excluded. Defining a goal to be achieved is actually defining a problem to be solved. An intelligence that is constrained by an already given goal or a set of well defined goals can hardly represent the ultimate generality of intelligence. In order to overcome this inherent partiality imposed by defined goals, suppose we could come up with a concept of a `universal goal' not dissimilar from the universal prior that generalizes the environment. Clearly, this will result in an absurdity since every actual sequence of actions the agent might come up with in order to achieve a subset of this universal goal, will be detrimental to another complementary subset. Defining a goal is a symmetry-breaking event that creates for the agent a \textit{unique perspective} regarding its relations with the environment. Only on the basis of such a perspective can the agent possibly operate intelligently. But again, similar to the formation of agent--environment distinctions, the determination of a perspective that brings forth clear goals does not necessarily happen all at once. It might well take place in a gradual and unique process of determination which involves a kind of intelligence that the above definition is entirely overlooking. The intelligent agent characterized in figure \ref{fig:Agent-Environment} is indeed a problem solver; still, it is argued that general intelligence never starts with solving a problem but much earlier -- in the formation or identification of the problems to be solved.  
	\item There is an unwarranted implicit asymmetry between the agent and the environment. While the agent is profoundly changeable by the environment i.e. it accumulates knowledge through learning and adaptation, the environment is only changeable within the limits of its givenness ( i.e. the actual yet unknown distribution of events it brings forth in the course of interacting). Such conceptualization excludes environments that are populated by other intelligent agents. The reason is that such intelligent agent(s), being part of the environment, may have a distribution of responses that cannot be determined or inferred in advance, at least not without some prior knowledge of their goals. In short, the definition in \ref{subsec:current_definition} does not consider cases of reflexivity where, for example, two (or more) agents interact without any \textit{a priori} knowledge of each others' goals and where such goals and their consequent behaviors emerge and consolidate in the course of interaction. If we consider that an agent's goals are set by an ongoing uniquely evolving perspective, it might be worthwhile considering an environment of co-evolving quasi-determined agents where the manifested intelligence profoundly departs from the presumptions made by the above definition\footnote{This is possibly what Goertzel means in his emphasis on complex environments in the definition.}.    	
\end{itemize}
In the light of these points of critique, it is clear that the currently accepted definition of general intelligence covers only a well characterized kind of intelligence but neglects the more profound and less easy to define process of the emergence of intelligence, or what we call \textit{open-ended intelligence}. The difficulty lies of course in the \textit{a priori} assumptions one is willing to give up. The less assumptions one commits to, the more difficult it is to make the concept concrete and formal. Wittgenstein famously said that whereof one cannot speak (clearly), thereof one must be silent. But then how can we explain babies learning to talk whereof initially nothing they say can be said to be clear? But still they do! Similarly, what is intelligence prior to anything intelligible? 

\section{The ontological roots of intelligence} \label{sec:ontological}
To try to answer this question, we need to reexamine a few deeply rooted axioms and explore the less charted conceptual grounds of how intelligence arises in the ontological sense. In other words, an attempt is made to reduce to the minimum the number of assumptions that constrain the concept. This is how we arrive at the concept of open-ended intelligence. As we will shortly show, it is a non-conventional concept; one that can never be spoken of clearly, but is not condemned to silence. The fluid and generative character of open-ended intelligence precedes and complements to the well established concept of intelligence that we criticize. Without such a complementary approach, it seems that a truly General AI is bound to remain beyond the reach of understanding. 

\subsection{The ontological ``chicken and egg'' problem} \label{subsec:chickenegg}
Much of how one thinks about anything including intelligence is already encoded in one or more of the major philosophical theories that shape human thought. These thought systems usually make explicit some set of ontological axioms of what is given prior to any thought or idea, and from there they proceed to derive all that can be thought or made sense of. Let us see how it works in the case of the definition of intelligence. The definition begins with a realist empiricist view that can be summarized in two seemingly simple assumptions:
\begin{description}
	\item[Realism] -- Posits that the whole of existence has an observer independent status. In our case it means that the environment exists independently of the agent interacting with it. Its structure and dynamics might be unknown to the agent but they nevertheless exist. Also, the agent's actions affect the environment only within the constraints of its independent givenness. 
	\item [Empiricism] -- Following Hume, posits that all sense-making and consequently all knowledge and intelligent behavior must derive from sense experience. In our case it means that for intelligence to manifest, it is necessary for the agent to interact/observe with its environment because only via interactions/observations can it learn what is necessary to achieve its goals \footnote{observations can be entirely passive but interactions are necessary in order to observe the effects of the agent's actions on the environment}.  
\end{description}
But already here there is a difficulty reminiscent of the chicken and egg problem: what comes first, experience or the subject of experience? In Deleuze's discussion of the human image of thought \citep[pp. 129-168]{deleuze_difference_1994}, the subject of sense experience, in our case the agent, cannot be an \textit{a priori} given as it is implied by the empiricist position. It must be somehow constituted in the course of sense-making. But this seems to be impossible because if we give up a subject \textit{a priori} to experience itself, who or what is there to experience in the first place? This is indeed the major point of Kant's critique of Hume's empiricism. The Kantian position necessitates a transcendental subject in possession of transcendental categories (such as space and time) antecedent to what is given in experience in order to make sense of experience. But Kant's approach is not without its own difficulties. It must assume that certain mental categories precede any actual thought and any manifestation of intelligence. It is like saying that some form of primal intelligence must be inherent in the agent prior to any interaction. But what would possibly be the origin of such primal intelligence that transcends experience? Clearly, the idea of general intelligence expressed by the definition in \ref{subsec:current_definition} follows Kant in assuming that the agent possesses certain capacities and goals prior to any observation or action. Here we face a second difficulty: how general is our agent's ``general intelligence'' if it must be constrained by \textit{a priori} categories that shape its observations and goals that assign values to them? Our thinking about intelligence seems therefore to be constrained by abstract patterns that shape conventional thinking itself. These patterns, collectively termed by Deleuze `the image of thought', draw implicit limits on intelligence itself.   

How can we overcome this difficulty and reach a conception of an open-ended intelligence? Following Deleuze \citep{deleuze_difference_1994,weinbaum_complexity_2014}, we should neither try to figure how the objects of experience produce subjects (Hume's empiricism), nor how the subjects of experience produce their objects (Kant's transcendental categories). Instead, Deleuze proposes what he calls \textit{transcendental empiricism}, a novel and seemingly paradoxical construction that affirms both Hume's and Kant's positions by redefining them. The position of transcendental empiricism starts with much fewer assumptions. It assumes neither subjects nor objects and instead of trying to figure how they might produce each other, it examines how both subjects and objects can be produced out of a field that initially does not assume either. Without delving more than necessary into the highly complex philosophical construction that is required here, we can start seeing where it leads in our case: giving up the \textit{a priori} givens in our thinking, namely, the agent, the environment, the distinction between them, the implied observations and actions that are made possible by such a distinction and finally the goal and its associated mapping of rewards. This might seem, at first sight, as if nothing is left to build upon and this clearly makes no sense. But here is exactly the point to stop and consider: if there is no sense, how is one to make sense out of a non-sense situation where no agents or objects can be identified to begin with? In other words, and here is the conceptual leap that needs to be taken, while the definition we started with in \ref{subsec:current_definition} is answering the question \textit{``what does it mean \textbf{to be} intelligent?''}, here the focus is on a prior question: \textit{``what does it mean \textbf{to become} intelligent?''}. Becoming intelligent is precisely this process of sense-making that precedes clear distinctions and goals and bring those forth. In order to see how is it possible at all, a novel and non-conventional set of concepts is required.             

\subsection{Individuals and individuation}
One of the most profound characteristics of the conventional system of thought humans use to make sense with is its focus on individuals. This focus has its roots in Greek philosophy and particularly in the metaphysics of Aristotle, which describes a world made of individual beings with an identity that is given as a set of stable properties and qualities. Aristotle's principle of the excluded middle ensures that an individual cannot possess a certain property while simultaneously not possess it. Hence, the identity of individuals, according to the Aristotelian theory, is unambiguously defined. 

Understanding the nature of individuals clarifies the general nature of definitions such as the one we use to define intelligence: definitions are made to delineate individuals. Most significantly, the focus on individuals also conditions the way one accounts for their genesis. To put it briefly, if individuals are the primary ontological elements of anything existing, the genesis of individuals is merely the manner by which one individual transitions into another one. Everything starts and ends with individuals while the becoming of individuals -- what happens in-between -- is secondary at best \citep{weinbaum_complexity_2014}. In order to make intelligence definable, therefore, we must make assumptions whose sole function is to comply with what the conventional system of thought dictates, namely, positing already formed individuals on the basis of which we can safely continue to develop further individual concepts and theories. 

Attempting to understand intelligence prior to such assumptions, we need a shift of perspective from individuals as the primary elements that occupy our investigation to \textit{how they come into being} in the first place, in other words, to their \textit{individuation}. Individuation is the formation or becoming of individuals. It is a primal formative activity whereas boundaries and distinctions arise without assuming any individual(s) that precede(s) them. The nature of distinctions and boundaries is subtle; inasmuch as they separate subject from object, figure from background, and one individual from another, they must also connect that which they separate. A boundary, therefore, is not only known by the separation it establishes but also by the interactions and relations it facilitates. 

This shift of perspective constitutes an alternative system of thought. Gilbert Simondon, the father of the theory of individuation \citep{simondon_individuation_2005} encourages us to understand the individual from the perspective of the process of individuation. For him, the individual is a metastable phase within a continuous process of transformation and is always impregnated with not yet actualized and not yet known potentialities of being:
\begin{quotation}
	``Individuation must therefore be thought of as a partial and relative resolution manifested in a system that contains latent potentials and harbors a certain incompatibility within itself, an incompatibility due at once to forces in tension as well as to the impossibility of interaction between terms of extremely disparate dimensions.'' \citep{simondon_genesis_1992}. 
\end{quotation}
According to Simondon, an individual is not anymore the rigid well defined Aristotelian element endowed with ultimately given properties, but rather a plastic entity, an ongoing becoming. The relatively stable state of individuals is punctuated by periods of transformation whereas individuals may radically change or disintegrate. Every such period reconfigures the inner tensions active within the individual and the manner by which they will determine future stable phases and transformations. 
\subsection{The condition of individuation}
Three descriptive terms stand out in Simondon's development of the concept of individuation: \textit{metastability, intensity and incompatibility}. These are in fact overlapping facets of the field of individuation. Imagine for example a system of two (or more) human agents in disagreement having an argument. As long as they both continue to engage with each other and haven't reached an agreement, the situation of their engagement is metastable. There are unrealized potentials of change in their relations. One of them may suddenly understand the other better and change her mind. Also the opposite can happen: the differences between them can grow and reach a point of crisis. The system may move both towards or away from stability in a manner which is not entirely predictable and depends on numerous factors. But as long as the argument continues, as long as the system is metastable, there is a motion of change. Individuation in this sense is reminiscent of the concept of self-organization in dynamic systems both in its reference to metastability and in the emergence of structure in a process of relaxing a system of tensions/potentials. But while self-organization commonly describes the convergence of trajectories towards attractors within an already configured state-space, individuation does not assume such an \textit{a priori} configuration. Simondon's notion of metastability is not confined to describing trajectories of movement among local minima within an already given landscape of potentials; metastability also involves possible transformations of the landscape itself (e.g. the number of the involved variables and the relations between them).  

Individuation takes place as long as the system has not reached a final stability/relaxation and exhausted all its potential for change. But in fact final stability is merely an idealization because it requires a closed system that either does not interact with its environment, or is  not distinct from its environment (i.e. in thermodynamic equilibrium with its environment). Open systems like living organisms or whole ecosystems maintain a far from equilibrium state \citep{prigogine_order_1984}, and are in a motion of continuous individuation never reaching permanent stability. 

The motion of individuation is driven by what can be called intensive differences, or in short, intensities. By intensity we mean here a general term for energetic differences that drive structural and state changes in a system (see, \citet[chap 2]{weinbaum_complexity_2014,delanda_intensive_2013}). In the example above, the driving intensities are the interlocutors' desire to each hold to her own convictions and persuade the other to change his. This desire is a force that drives and animates the interaction. Intensities can either dissipate as the system changes, or they can also become too strong for the system to contain and thus bring about the disintegration of the system. Applied to our example, in both cases the activity of arguing will tend to cease. If the interlocutors manage to agree on a certain point, intensities are relaxed and their relations gain additional consensual structure (understanding). But if, on the contrary, they discover that their differences are even deeper than they initially thought, intensities increase and may find their expression in the manner the argument is conducted e.g. it becomes heated, or even escalates to physical violence, which is not anymore an argument. Generally, intensities are correlated to the measure of metastability and level of structural changes taking place in the system. Low intensities are associated with relatively more stable dynamics while high intensities are associated with volatile dynamics and swift structural changes. 

Last but not least is the third term -- incompatibility. Only situations of incompatibility bring forth intensities that drive processes of individuation. Incompatibility arises from what we may call the problematic -- the situation where interacting elements of a system pose problems to each other that require resolution. The engagement of predator and prey is an exemplar of a problematic situation of incompatibility. In the argument example above, thinking differently about a situation that requires from the agents a joint coordinated action is an example of a problematic situation. The differences in perspective between the agents must be resolved at least to a degree that allows the necessary joint action. Disparity is an extreme case of the problematic where the semantics of the signs exchanged between agents/elements in a system is not established or ambiguous. The agents lack a common ground of basic coordination/understanding to even facilitate their engagement (e.g. they do not speak the same language). In such cases, individuation must also mean the emergence of a coordinated exchange of signals (is this strong hug a gesture of friendship or a covert threat?). It is important to note that the individuation of systems in general always starts from a situation of disparity. It takes place in the course of gradually establishing a coordinated exchange of signals among gradually differentiating elements that together (distinct signals and elements)bring forth a system. In other words, both elements and the relations among them are simultaneously individuated. Furthermore, individuation never brings forth an individual in a vacuum but rather an individual-milieu dyad. This dyad contains both a system of distinctions and a system of relations. The individual and its milieu reciprocally determine each other as they develop as a system greater than the individual \citep{simondon_position_2009}.

\subsection{Transduction -- the mechanism of individuation} \label{subsec:transduction}
What happens in individuation? In the example of arguing persons, the involved agents are continuously affecting and being affected by each other. In the course of their interactions some (but not necessarily all) of the disparities and problems are resolved and result in a new consensual structure that they will support together in the future. This is how the system is individuated and gains an identity of its own based on the established coherency achieved between the agents. It is important to note that at any instance the  system constituted of the agents and their relations includes both consensual positions that form its individuated aspect (because they can be identified and defined for the entire system), and elements of unresolved incompatibility that may drive future engagements leading either to extended coherency or the destabilization of the already established consensus. What may seem to an external observer as a stable and coherent system, always harbors internal intensities and instabilities that threaten to radically change it or even break it apart. These latter elements, termed \textit{preindividual}, are intrinsic to all individuals and are the inner intensities that drive future individuation.

The outcome of the interaction between two or more incompatible agents is hardly predictable since it is not guided by \textit{a priori} individuated overarching principles or mechanisms. In other words, the outcomes of such interactions can neither be deduced from an already individuated setup, nor can they be induced from a generalized model based on previous similar instances because incompatibilities are inherently singular and unrepeatable. The methods of deduction and induction therefore cannot be applied to individuation \citep[p 12]{simondon_position_2009}. Prior to, and in the course of the actual interaction, the outcome is said to be \textit{determinable but not yet determined}. Determination necessitates the actual localized and contextualized interaction where the participating elements reciprocally determine behavioral and structural aspects of each other. This kind of interactions constitutes the mechanism by which individuation takes place as a sequence of \textit{progressive determinations} and is called in short \textit{transduction}. Transduction is an abstract mechanism that may receive its specific actual description per context or operational domain. It can be physical,biological, cognitive, social or other according to the agents and interactions involved. 

The most important aspect revealed in transduction is the progressive co-determination of structure and behavior. Transduction can be seen as a chain of operations $O_i$ on structures $S_j$: $S_1 \rightarrow O_1 \rightarrow S_2 \rightarrow O_2 \rightarrow S_3 \rightarrow$... \citep[pp. 14-15]{combes_gilbert_2013}. Every operation is a conversion of one structure into another, while every structure mediates between one operation and another. Each structure in the chain constrains the operations that can immediately follow. Each operation, in its turn, can transform the previous structure into a limited number of new structures. Every intermediate structure is a partial resolution of incompatibility but it is driven away from its relative stability as long as the remaining unresolved tensions are not exhausted. Initially, the series of operations or structures can be quite random. As the transductive process progresses, certain structures and operations may become more frequent than others or even repetitive. As the sets of structures and operations become mutually bounded, even temporarily, an individuated entity arises which may either further consolidate or eventually disintegrate (this is further discussed in \ref{subsec:phases}). A more concrete example of a transductive process is the propagation of a computation in self-transforming programs: executed code and the data are analogous to operation and structure. However, the program code itself is also accessible as data that is progressively modified to produce (in principle) inexhaustible variety and innovation. Code redefines the data and data further redefines the code in a chain of operations. The analogy helps to understand how operation and structure are reciprocally determining expressions of the transductive process. 

\subsection{Assemblages - from Individuation to Individuals} \label{subsec: assemblages}
Understanding individuation, is understanding how individuals are constructed from sets or populations of disparate and heterogeneous elements. The monolithic stable character of individuals is given up and instead we see metastable and often troublesome constructions that can be defined and identified but only as provisional stations in an incessant process of transformation. These constructions are called \textit{assemblages} -- a concept developed by by \citet{deleuze_thousand_1987} and further extended by \citet{de_landa_new_2006}\footnote{A short introduction to assemblage theory can be found in:\\ \href{http://wikis.la.utexas.edu/theory/page/assemblage-theory}{http://wikis.la.utexas.edu/theory/page/assemblage-theory}.}. Assemblages are networks of interacting heterogeneous individuals where each individual is an assemblage too (for elaboration on the stratification of assemblages see \ref{subsec:descriptive}). Assemblages carry with them an intrinsic though metastable individuality; an individuality that does not depend on an external observer but only on the relations that have been stabilized among their disparate elements. 

\label{page:knowledge} When we observe a system of any kind, be it a physical object, an organism, a technological artifact or a social system, as observers we form with the object of observation a new assemblage. Both observer and observed and the relations between them undergo a transductive process of individuation where disparities are resolved and coherent relations are established. These come to constitute knowledge -- an individuated knowledge. The individuation of knowledge is the process already mentioned in subsection \ref{subsec:chickenegg} where both the subject and object of knowledge reciprocally determine each other without one having a privileged ontological status over the other.  

Replacing the individual with individuation as the primary ontological element exposes a hierarchy of creative processes across many scales. When we examine an individual, we need to identify which are the elements relevant to its individuation. For example, to say that a living organism is made of atoms does not expose anything interesting about the organism's individuation. An individual organism individuates from a lump of identical cells originating from a single cell in a developmental context (egg, womb, or a cell membrane in case of unicellular organisms). An individual species individuates in an evolutionary context (ecology). In the case of social animals, further individuation takes place in a social context. Every such context can be given as a population of heterogeneous and disparate elements from which individuals and their milieu co-emerge.  

Individuals as assemblages comes to mean that the assembled elements can be said to be characterized by 
\begin{inparaenum}[\itshape a\upshape )]
\item identifying properties that define them as the individuals that they are (and subject therefore to their own individuation) and 
\item capacities to interact -- to affect and be affected by other elements \citep[chap 1]{de_landa_new_2006}.	
\end{inparaenum}
While every element has a more or less fixed and independent set of properties, the set of its interactive capacities is open and inexhaustible because it depends on the actual relations that each element forms with other elements. Since there is no limit to the number and kind of relations, the set of capacities to interact is open-ended and non-deterministic. What becomes determined in transduction are the actual capacities manifested in the interactions of the various elements involved in the process. This is why the actual interaction is necessary for the determination and why the resulting relations cannot be predicted \textit{a priori}. Once disparate elements come into (partially) coordinated relations they give rise to an assemblage -- a new individual with more or less stable properties and capacities\footnote{Capacities also mean capacities to be destroyed. Certain interactions can amplify the internal unsettled intensities within an individual to the effect of its disintegration.}. Critical to the concept of assemblage is the semi-autonomous status of their constituting elements and their contingent relations. Even in a radical example such as an organism or even a living cell where the integration between the constituting elements seems to be very tight, the relations between organs are not a result of a logical necessity but rather contingently obligatory \citep[p 12]{de_landa_new_2006}. This is why tissues can spontaneously become cancerous and individual organs can be taken out and replaced by artificial organs such as bionic limbs, artificial kidneys, hearts, joints, retinas etc. to form cyborgian assemblages.

To summarize, transduction and assemblage are both aspects of individuation. While transduction describes the dynamical aspect, assemblages frame the structural aspect. Together they form a conceptual framework that allows the investigation of the individuation of intelligent agents free of the assumptions reviewed in subsection \ref{subsec: criticism}. An interesting reflection arises from the distinction between properties and capacities: since the conventional approach to general intelligence conceives it as a definable property of an agent or a system, there is a certain inherent finiteness in its very conception. In contrast, the complementary approach to general intelligence proposed here conceives general intelligence as a capacity. As such, it is open-ended and involves indefinite and \textit{a priori} unknown factors depending on contingent interactions. In other words, a creative aspect that goes beyond goal-oriented, utility-optimizing activity, is intrinsic to the open-ended intelligence that manifests itself in individuation.

\section{Intelligence in the individuation of cognition} \label{sec:individuation_of_cognition}
From the perspective developed here, it is interesting to regard every individual -- the product of individuation -- as a solution to a problem whose formulation is not initially given and can only be implied 
from the solution (i.e. the individual). As was already mentioned, individuation is a resolution of a problematic situation by means of progressive determination. What is meant by ``a problematic situation'' is a state of affairs that is unstable, non-organized, whose elements lack definite boundaries and coordinated interactions and therefore does not give itself to any systematic description.

In contrast to the definition given in \ref{subsec:current_definition}, where the starting point is a well formed problem to which a solution is being sought, in individuation the endpoint i.e. the individual, is a well formed solution to a `problem' that is initially unformed and therefore can only be implied `backwards' from the solution. Following this line of thought, we do not depart in any radical sense from the understanding of intelligence as a general capacity of solving problems. Our thesis is that the formative processes that bring forth individuals as `solutions' to problems that are initially unformed, are manifesting an open-ended kind of intelligence which is qualitatively different and complementary to the one defined in \ref{subsec:current_definition}. 

Furthermore, the designator `general' in general intelligence must relate to a process where a problematic situation is initially unformed (i.e. determinable but not determined) but involves intensities that mobilize the bringing forth of individuals as `solutions'. The most significant example that comes to mind  here is the process of natural evolution. Every living organism manifests a set of behaviors that realize highly optimized solutions to problems that become apparent only while observing the interactions of the organism in its environment. Organisms therefore are undeniably intelligent. But what about the evolutionary formative processes that give rise to the outstanding `solutions' that living organisms are? We argue that they are intelligent precisely in the complementary sense we propose here and in this profound sense deserve, more than anything else, the designation of General Intelligence. 

To this point, the theory of individuation has a very broad scope as it relates to individuals in general and not necessarily to what is conventionally considered intelligent systems or intelligent agents. From a  philosophical perspective, general systems whether natural like galaxies, stars, rivers, chemical compounds, weather systems etc., or artificial such as tools, machines, buildings, wars, mathematical computations etc. are individuals in the course of individuation that possess an intrinsic and identifiable (though in most cases limited) manifestation of intelligence. Our interest however is not in the limited and already consolidated manifestations of intelligence but in those manifestations which are, at least in principle, open-ended i.e. in the process of becoming intelligent. In the following we develop the idea that the individuation of cognitive systems, where cognition is understood in its broadest possible sense, is by definition a process manifesting open-ended intelligence.  

\subsection{Cognition as sense-making} \label{subsec:enactive}
The phenomenon of cognition is definitely complex, multifaceted and gives itself to quite a few diverse definitions. Still, in a somewhat naive approach, the activity of cognition is naturally associated with certain situations when there is an agent operating in its environment, and whose operation can be described as an ongoing problem-solving activity. In other words, the roots of cognition is always a problematic situation, an incompatibility, full of tensions, that exists between the agent and its environment and that needs resolution somehow. This also lends the impression that a cognitive agent is always involved in some purposeful activity, that is, resolving an immediate (or a forethought) problem. This is also the straight-forward manner by which cognition is associated with intelligence. 

Here, we turn full circle to the beginning of our inquiry: how is it that this setup of agents, environments and their dynamic problematic relations emerge in the first place? Even while writing (or reading) these words, we make use of sensible objects that are already, at least partially, formed. Perhaps they are vague and require further determination to become clearer; some may change the meaning (sense) in which they are understood; others may just emerge in the flow of thought or disappear; and yet others may merge or diverge. Crossing this, often unseen, boundary between the unknown and the known, the unformed and the formed is what we may call \textit{sense-making}. Sense-making is the bringing forth of a world of distinctions, objects and entities and the relations among them. Even primary distinctions such as `objective -- subjective' or `physical -- mental' are part of sense-making.           

A relatively new appearance on the stage of cognitive science, the so called \textit{enactive cognition} approach, regards sense-making as the primary activity of cognition. The term `enactive', synonymous with `actively bringing forth', makes its first appearance in the context of cognition in the book ``The embodied Mind''  \citep{varela_embodied_1992} and has been since then the subject of many developments and debates \citep{stewart_enaction:_2010,thompson_mind_2007,di_paolo_autopoiesis_2006,de_jaegher_participatory_2007}. 
The enactive theory of cognition incorporates the idea of individuation rather naturally as it asserts cognition to be an ongoing formative process, sensible and meaningful (value related), taking place in the co-determining interactions of agent and environment \citep{di_paolo_horizons_2010}. Still, being based on the earlier works of Maturana and Varela on autopoiesis and the biological basis of cognition \citep{maturana_tree_1987,maturana_autopoiesis_1980}, the theory asserts the necessity of an autonomous and relatively stable identity to cognition: 
\begin{quotation} \label{enaction}
A guiding idea of the enactive approach is that any adequate account of how the body can either be or instantiate a cognitive system must take account of this fact that  the body is self-individuating. This point brings us to the principal concept that differentiates enactivism from other embodied approaches to the mind - the concept of autonomy. \citep{di_paolo_enactive_2014}
\end{quotation} And in \citep{di_paolo_horizons_2010} the necessity is made specific to sense-making:
\begin{quotation}
[...] By saying that a system is self-constituted, we mean that its dynamics generate and sustain an identity. An identity is generated whenever a precarious network of dynamical processes becomes operationally closed. [...]  Already implied in the notion of interactive autonomy is the realization that organisms cast a web of significance on their world. [...] This establishes a perspective on the world  with its own normativity[.] [...] Exchanges with the world are thus inherently  significant for the agent, and this is the definitional property of a cognitive system: the creation and appreciation of meaning or  sense-making, in short. [...] [S]ense-making is, at its root, the evaluation of the consequences of interaction for the conservation of an identity. \citep[pp. 38-39,45]{di_paolo_horizons_2010}	
\end{quotation}
In contrast, we argue that the broader understanding of sense-making as the individuation of cognition itself, precedes the existence of already individuated autonomous identities and is actually a necessary condition to their becoming. Only that at this pre-individuated stage there is still no one for whom sense is being made. It is only a habit of thought to assume the preexistence of the sense-making-agent to the sensible (see \ref{subsec:chickenegg}). Di Paolo et al. are nevertheless aware of the metastability involved in the processes that constitute cognition by mentioning \textit{precarious networks of dynamic processes becoming operationally closed}, but they do miss the deeper meaning of becoming as a process and therefore treat closure as an ideal point that delineates the existence of the individual in time, and that only from such a point and on sense-making is made possible. This is an important point because it frees intelligence from being conceptually subjugated to the persistence of a preexisting identity. The sensible, we argue, precedes the individual and facilitates its becoming but in itself is not necessarily biased towards the conservation of any identity.
In sense-making, both integration and disintegration play a significant role. 

To summarize, the manifestation of open-ended intelligence in cognition is the bringing forth of a complex world via the activity of sense-making. The concept of sense-making captures two distinct meanings: the first is synonymous with cognition as a concrete capacity, the second, with the individuation of cognition as intrinsic to cognition itself. The latter meaning of sense-making is the one corresponding to the acquisition and expansion of concrete cognitive capacities (i.e. intelligence expansion) and it also generalizes the concept of \textit{cognitive development} beyond its psychological context \citep{piaget_principles_2013}.

\subsection{A descriptive model of the individuation of cognition} \label{subsec:descriptive}
To describe the process of individuation of cognition in more concrete terms, we consider a heterogeneous and diverse population $P$ of individual elements each with its defining properties and capacities to affect and be affected that depend on contingent interactions with other elements of the population. By `heterogeneous' we mean a population of individuals with different sets of properties whereas by `diverse' we mean that there is variability in the expressions of at least some of the properties. An obvious example would be a population of organisms within an ecology: The population is heterogeneous because there are many species and it is diverse because specific properties have variability in expression within a specie and across species. The formation of new individuals within heterogeneous and diverse populations of interacting elements is at the core of our model. It highlights the distributed nature of individuation and the kind of intelligence that is thus brought forth.

As already described in \ref{subsec: assemblages}, individuals are actually individuated assemblages. For reasons that will become clear shortly, we assign the population we start with to a stratum $P$. Stratum $P$ implies two additional populations (i.e. strata) with which it holds hierarchical relations:
\begin{enumerate}
	\item Lower in the hierarchy is the population of all the individuals that participate as components in assemblages that constitute the individuals in stratum $P$. We mark it $P_{sub}$ for being the substratum of $P$.
	\item Higher in the hierarchy is the population of all the individuals whose assemblages are constituted from individuals in stratum $P$. We mark it $P_{sup}$ for being a superstratum of $P$. 
\end{enumerate}

\begin{figure}[h]
	\centering
	\includegraphics[scale=0.19]{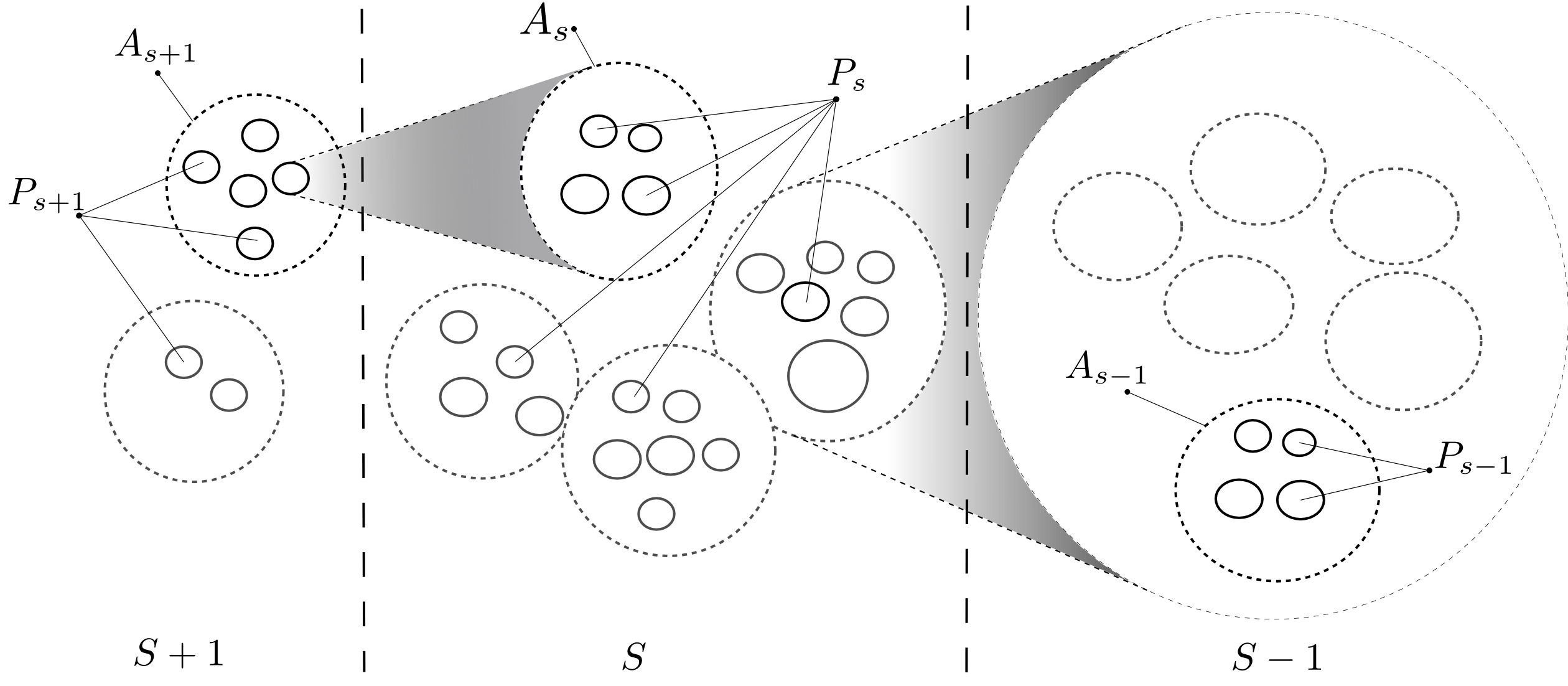}
  	\caption{Relationship between strata in the model: $S$ consists of $P$, $S+1$ consisting of $P_{sup}$ is the superstratum and $S-1$ consisting of $P_{sub}$ is the substratum. $P_{s}$ denotes the population of agents at stratum $S$. Solid circles denote the individual agents at any stratum. Dashed lined circles denote assemblages at any stratum e.g. -- $A_{s}$ at the center of the figure, denotes a super-agent that emerges from the interactions of agents in $S$. Assemblages at stratum $S$ are the individuated agents of stratum $S+1$.}
  	\label{fig:illustration_of_scales}
\end{figure} 

This hierarchical relation of assemblages unfolds recursively both upwards and downwards where each level is the substratum of the level above it. Lower levels are populated by successively simpler elements and higher levels are populated by successively more complex elements so different levels in this hierarchy are of a different scale of complexity\footnote{The scaling is not only structural but also temporal. Changes at different scales do not occur at the same frequency, also the stability of elements varies with complexity.}. 

This simple scheme allows us to describe the transductive mechanism operating at stratum $P$ from two distinct perspectives:
\begin{inparaenum}[\itshape a\upshape )]
	\item $P_{sub}$ provides the `raw material' perspective for the processes in $P$ in terms of already individuated elements that are given\footnote{The designation `given' here is a simplification made for clarity. In fact the elements in $P_{sub}$ are never fully individuated and are affected by interactions taking place in its two adjacent strata as well.} and
	\item $P_{sup}$ provides the `product' perspective for the processes in $P$ in terms of the individual objects that are individuated in $P$.
\end{inparaenum}
Stratum $P$, therefore, is a field of individuation where individual elements from $P_{sub}$ get assembled
by the actual interactions taking place in $P$ to produce the higher level individual elements in $P_{sup}$.
The assemblages that emerge in $P$ are products of a sense-making process taking place in $P$ and therefore can be said to become \textit{sensible} in $P$.   

The individuals operating at each strata can be broadly defined as agents considering their capacities to affect and be affected. Specifically, in our model, the individuals described at each stratum are cognitive agents whereas their capacities grow in complexity across strata from the most primitive distinctions and actions at lower strata to highly complex sense-making activities at the higher strata. Even so, at this level of description, we do not have to assume agents with intrinsic values or goals that require a certain level of autonomy as discussed above in \ref{subsec:enactive}. We only need to require that minimally some of the capacities to affect are also within what is possibly affected by interactions with other agents. In other words, the agents' behaviors are, at least to a minimal extent, affected by their interactions. 
This requirement comes to ensure that some individuation can take place. Clearly, if the agents' capacities to affect had been entirely independent from their capacities to be affected (and vice-versa), they could not change one in relation to the other and therefore no transductive process would have been possible in such case.   

In summary, we introduce two distinct kinds of relation among agents; horizontal and vertical. Horizontal relations are internal to each stratum and describe the actual interactions that bring forth individuation. The vertical relations are across adjacent strata and describe hierarchies of individual objects differing in complexity and their upward and downward effects. While conceptually the individuation of elements at any stratum follows the same transductive mechanism, the actual mechanisms are context dependent and can be vastly different; the resolution of disparities between neurons, for example, is nothing like the resolution of disparities between goals, needs and constraints in the mind of a single human individual, and is nothing like the resolution of disparities between humans or between social organizations constituted of humans and their artifacts. Nevertheless, the guiding principle of individuation and its self-similarity across strata introduces a general model of cognition that is scalable and open-ended.  

There is no end, in principle, to the possible expansion of intelligence via the emergence of new strata of individuation. Whenever a population of cognitive functions/objects emerges with enough diversity and heterogeneity to become the substratum of novel individuations, a higher stratum of cognition can potentially emerge. The emergence of a population of a new kind of individual (e.g. new species in macro evolution, a new kind of explanations allowed by a novel theory, mobile devices, applications of deep learning algorithms etc.) can be thought of as a phase transition event in sense-making where certain kinds of assemblage that were rarely present before, if at all, suddenly become ubiquitous, diverse and heterogeneous. When such an event takes place it can often be associated with a new method or set of methods of resolving problematic situations and coordinating elements into wholes that could not be integrated before. Phase transitions in sense-making is possibly the underlying driving principle behind \textit{metasystem-transitions} -- a theory of the evolution of complexity in general systems \citep{turchin_phenomenon_1977,turchin_dialogue_1995,heylighen_evolutionary_2000}.

\subsection{Phases of sense-making} \label{subsec:phases}
Actual sense-making is a continuous process of integration and disintegration of discrete individuals taking place in a network of agents and their interactions. In the context of cognition, sense-making is synonymous with individuation. It is important to note that in our general approach to cognition there is no \textit{a priori} subject who `makes sense'. Both subjects and objects, agents and their environments co-emerge in the course of sense-making. For clarity of description, a few phases can be identified in the process, given that this deconstruction into phases is largely didactic. 
\subsubsection*{Preindividual boundary formation}
The spontaneous emergence of an agent-environment dyad from a random network of interactions can be thought of as the formation of a boundary that distinguishes a subset of agents in population $P$ from all the rest. Boundary formation corresponds to self-organization in the broadest sense. Once there is a boundary, interactions across the boundary also gain a distinctive significance in the sense that now the set of all possible interactions can be further categorized in relation to the boundary, i.e. those interactions across the boundary and all the rest. Therefore, the formation of a boundary is equivalent to symmetry breaking over the population of agents and their interactions. 

Initially, boundaries that arise are not fixated and possess no tendency to persist. Nevertheless boundaries can persist, for a while, even without actively resisting change if they are not perturbed. How is the spontaneous formation of boundaries possible? Without specifying the exact nature of the interactions that are responsible for that, we can assume  that in a network of interacting agents, there  will exist a non-uniform distribution of interactions over the population. There will be subsets of agents that spontaneously affect each other more strongly or frequently than they are affected by the rest of the agents of the population. Observing such a network for long enough and drawing a map of the density of interactions, one would, in most cases, find regions of higher density of affective interactions (i.e. interactions that change the state of the participating agents) compared to their surroundings. This non-uniformity of affective interactions can be further quantified in information theoretic terms following the concept of \textit{information integration} developed by Guilio Tononi \citep{tononi_information_2004,tononi_consciousness_2008,edelman_universe_2000} in the context of computational neuroscience. A simplified mathematical development of the concept can be found in \nameref{app:appendixA}. 

The information integration of a set of interacting agents is a relative measure of how strongly their states have become mutually correlated in comparison to their correlation with the rest of the environment. In our case, information integration is used as a clustering criterion that singles out from the population $P$ subsets of agents that are significantly more integrated in the sense of affecting each others' states. The information integration of groups of agents requires no \textit{a priori} assumptions regarding their dynamics. In other words, informational integrated clusters can contingently arise and spontaneously persist for a while. But this contingent arising is sufficient to initiate a process of individuation eventually bringing forth order out of disorder.

Information integration is a necessary indication to boundary formation but is not a sufficient condition to individuation. What seems to be necessary for boundaries to consolidate and persist is an additional element of regularity or repetition in the interactions. This element is perhaps best reflected in Deleuze's introduction to the English translation of his book on Hume's empiricism:
\begin{quotation}
We start with atomic parts, but these atomic parts have transitions, passages, ``tendencies'', which circulate from one to another. These tendencies give rise to habits. Isn't this the answer to the question ``what are we?'' We are habits, nothing but habits[.]\citep[p. x]{deleuze_empiricism_1991}
\end{quotation}
At this very primal phase of boundary formation we are interested in what would be minimally sufficient to make contingent boundaries more distinct and persistent and by that drive individuation further. In our model, interactions take place with some initial probability distribution, and this leads to an initial distribution of information integration. The missing element can be understood as a kind of a cybernetic mechanism that reinforces differences in information integration, that is, integrated clusters of agents will tend to increase the probability of future interactions (and subsequent correlations in state) within the cluster in proportion to the present degree of integration. In other words, similar to Hebbian learning \citep{hebb_organization_1968}, agents that are already correlated to some extent will tend to increase their correlating interactions whereas the frequency of other interactions will relatively decrease.    

The reinforcement of the frequency of interactions is a general and simple conceptualization of the `habit forming' tendency. With such a tendency, not only do boundaries form spontaneously but they will tend to become more distinct once formed. Various specific reinforcement mechanisms are possible; the reinforcement can depend (positively or negatively) on the kind of interactions and the content of the information being exchanged between agents, but at the moment, we are only interested in the conceptual framing of a model of individuation with minimal assumptions. With this additional cybernetic element, the activity taking place within the network at any moment $T$ influences the future structure of the network at times $t>T$ as it makes certain links stronger than others. Also, the interacting agents gradually co-determine their future interactions. These two effects are the definitional marks of a transductive process going on, as we have seen in \ref{subsec:transduction}.

\subsubsection*{Closures,autonomy and identity} \label{subsec:closure}
The phase of sense-making that corresponds to already formed individuals is characterized by the emergence of special types of dynamic structures called \textit{operational closures}. Operational closure is a central concept in the enactive approach to cognition and is the basis of the so-called self-constituted or autonomous systems with identity that were already mentioned in \ref{subsec:enactive}:
	\begin{quotation}
An identity is generated whenever a precarious network of dynamical processes becomes operationally closed. A system is operationally closed if, for any given process $P$ that forms part of the system (1) we can find among its enabling conditions other processes that make up the system, and (2) we can find other processes in the system that depend on $P$. This means that at some level of description, the conditions that sustain any given process in such a network always include those conditions provided by the operation of the other processes in the network, and that the result of their global activity is an identifiable unity in the same domain or level of description (it does not, of course, mean that the system is  isolated from interactions with the environment). Autonomy as operational closure is intended to describe self-generated identities at many  possible levels \citep[p. 38]{di_paolo_horizons_2010}.		
	\end{quotation}
Implicit in this definition are a few important points. First, certain capacities of the agents (processes) involved gain significance as they become enabling conditions to the operation of other agents. The generality of affecting and being affected is further determined here because it specifies \textit{how} certain agents affect or are affected by others. Conceptually this implies a certain level of compatibility among the agents involved and therefore it means that for operational closures to arise, certain compatibilities among the participating agents must be present. These compatibilities provide a common descriptive ground that allows the various heterogeneous agents and their interactions to be described, at least in part, within the same level of description. Second, closures imply the existence of closed loops of interactions (i.e. topological determinations) among the participating agents and additionally the recurrence of certain sequences of specific interactions (i.e. behavioral determinations). Third, the use of the term `precarious network' hints that the autonomous construction is pretty fragile. If even one of its constituent agents does not fulfill its function, the whole construct might disintegrate. At least we can expect a significant and abrupt modification of identity in such cases. But the precariousness aspect is essential for the enactive approach as it ensures that the preservation of identity must somehow be an activity and not merely an inert property of the system. This is how a cognitive system is distinguished (see p. \pageref{enaction}).

Undoubtedly operational closures with distinct intrinsic characteristics and that ``follow laws set up by their own activity''\citep[p. 37]{di_paolo_horizons_2010} are what we normally consider as individuals. The continuity of self-generated identity becomes the basis of a \textit{perspective} an autonomous system has on its environment and a unique principle of sense-making subjugated to that identity and its persistence as a prime directive. Interactions across and within the boundary gain relevance and value in relation to this directive. But once the concept of identity and its continuity take root, individuation seems to have reached its end as the autonomous system will tend to resist further changes, or in other words, to exhaust its metastablity and reach a stable regime of its dynamics where it can regulate its interactions with the environment. To somewhat soften this apparent rigidity of autonomous constructs \citet{di_paolo_autopoiesis_2006} proposes what he calls a system's \textit{viability set} as the set of external perturbations and internal structural changes an autonomous identity can withstand without disintegration. We take this idea a definitive step further. 

Can there be a third phase of sense-making that incorporates both dynamic boundary formation and operational closures? We argue that not only does such a phase exist but that it is the case in the majority of actual phenomena. More often than not individuals are not rigidly fixed, but rather have continuously individuating  \textit{fluid identities}. Specifically, cognition as the activity of sense-making is never a stable set of competences that have exhausted all its potential for transformation, but is rather undergoing a continuous process of development\footnote{Development generally means increase in intelligence in correlation to the complexity of situations and objects the system can make sense of. But the process is not necessarily  monotonous; disintegration of already integrated structures can take place as well in the course of development. For example, when a theory is being replaced by a different, better theory that can explain and cohere more observations.}. 

\subsubsection*{Fluid identities} \label{subsec:fluid}
The idea of fluid identities is a modification of the enactive approach to cognition based on replacing individuals with individuation. The requirement of precariousness at the basis of an autonomous structure can be relaxed in the following important manner: that operational closures need to be maintained continuously means that critically the very property of closure is maintained but it does not necessarily mean that \textit{it is exactly the same closure that is maintained}. A closure $C$ can be maintained as a series of individual closures $C_1,C_2,C_3,...,C_i,...$ that share among them some or most of their constituent agents but still significantly differ from each other. The precariousness can therefore be said to be maintained as a global  property but is not locally maintained. The ordered set $[C_i]$ as a whole is then considered an individuating object with a fluid identity in the sense that it preserves most (but not all) of its invariant operational properties across short periods of time (e.g. while changing from $C_i$ to $C_{i+1}$), but there is also a slow drift of these properties such that after a long time and many consecutive transformations (e.g. changing from $C_i$ to $C_{i+k}, k\gg1$) the said object has possibly become radically different from how it began. How is this possible? Conceptually, we already saw in \ref{subsec: assemblages} that individuals are assemblages whose constituting components are themselves independent individuals that affect and are affected by each other. Components can be plugged into and out of the assemblage without losing their individuality (because their individuality does not depend on the interactions but rather on their intrinsic properties). Fluid identity is in fact the only proper description of an assemblage or a continuously individuating agent. They may lose or gain components in the course of their interactions. Some of these interactions may bring forth operational closures that did not exist before, others may disrupt already existing closures, and yet others may only replace one configuration of closure with another, possibly causing temporary but not fatal gaps in existing closures. All these movements are possible within assemblages and do actually happen all the time all around. That we tend to see the world in terms of stable identities is only an habit. Stable identities arising from strict operational closures are special cases of fluid identities where an assemblage has become (almost) crystallized or is just changing very slowly compared to its surroundings.  \vspace{10pt}  	

The phases of sense-making, from preindividual boundary formation, through fluid identities to autonomous closures, form in fact a single continuum of change that spans from ultimate disparity (disorder) to highly organized cognitive agents. Enclosures defined by information integration are preindividual and are characterized by a majority of contingent interactions over coordinated ones. Enclosures defined by operational closures are capable of sustaining an identity and are characterized by a majority of coordinated interactions over contingent ones. 

On the thick borderline between these, exist fluid identities that are manifestations of more or less balanced proportions between coordinated and contingent interactions. These are volatile entities whose defining characteristics change across time. These may radically change their closure construct or even temporarily lose the strict closure property altogether without losing their overall distinctiveness from their environment in the long run. From the perspective of open-ended intelligence these are the more interesting situations where new sense objects may arise out of no-sense but in association with previously established sense objects. This borderline seem to be where intelligence expands.    

\subsection{The resolution of disparity} \label{subsec:resolution}
As we have already mentioned earlier, the nature of intelligence intrinsic to individuation processes is associated with the resolution of disparity and problematic situations in a population of interacting agents, i.e. achieving higher degrees of compatibility. Compatibility is a general concept that distinguishes between ordered and disordered relations, structural, dynamic or both within the population of agents. Two agents are incompatible or disparate if their behaviors are entirely independent from each other. In interactions taking place between disparate agents, each will present for the other a source of unintelligible noise. No correlated or coordinated exchange of signals takes place in such a case. Consequently, the behavior of one agent cannot be inferred from observing the other. Collections of disparate agents do not constitute systems as yet. They require an exhaustive description of all the unique agents and behaviors. A system arises from a collection of agents only when some degree of compatibility is achieved between its member elements. Systems can have a more compact compressed descriptions (relative to their disparate initial state) because compatibility means a degree of regularity, similarity and recursion in structure and dynamics. The integration function $I(P)$ defined by equation \ref{eq:integration} in \nameref{app:appendixA} can be considered as a simplified general measure of compatibility. 

But compatibility thus understood cannot be the only factor necessary to qualify intelligence. A system with a highly compressed description would mean that its components are so highly compatible that it becomes redundant in terms of its properties and capacities ($I(P)$, accordingly, will be large). We need therefore to define a second factor we call \textit{operational complexity}. Qualitatively, the operational complexity of a population $P$ of interacting agents is the degree to which the overall system's states are differentiated. In other words, how many distinct behaviors it can present. A simplified measure of operational complexity $OC(P)$ can be given in information theoretic terms and is developed in \nameref{app:appendixB}. 

Clearly, a disparate collection of agents achieves the highest operational complexity since the states of all the agents are independent. But this extreme situation is actually not very intelligent (i.e. it is stupid). As each element operates on its own the emergence of collectively integrated informational states is impossible. In terms of sense-making, ultimate disparity indicates no boundary formation at all while ultimate integration indicates a redundant object with few or no inner states (i.e. no interesting behavior). A measure of the intelligence embedded in the dynamics of an assemblage of interacting agents must therefore consider a balanced combination of both information integration and operational complexity. Based on the mathematical derivations in appendices A and B, a measure of the open-ended intelligence operating in $P$ can be expressed as a function of both the compatibility measure $I(P)$ and the operational complexity $OC(P)$:
\begin{equation}
	Int_{t}(P)=\mathcal{F}(I_{t}(P), OC_{t}(P)) 
\end{equation}
The subscript $t$ here indicates that this measure is time dependent. It changes in the course of individuation and does not necessarily achieve maximal values in relatively stable individuals. This conceptual formula helps to establish that the resolution of disparities and problematic situations is not captured only by achieving compatibility between the disparate components. The open-ended intelligence intrinsic to the formation of an assemblage is correlated to both its inner compatibility and operational complexity. Compatibility only reflects a degree of integration existing in a collection of interacting agents; it does not indicate \textit{how} such integration is achieved. In order to resolve disparity and achieve compatibility, agents must \textit{coordinate} their interactions. \textit{Open-ended intelligence in individuating processes can therefore be associated with the coordination achieved by initially distributed disparate agents in the course of their interactions}. Coordination is what happens among agents that affect each other in a non-random manner but still maintain a significant degree of distinctiveness in their milieu. Whereas distinctiveness here means that an agent's behavior is not redundant and cannot be entirely given in terms of other agents' behaviors. $Int_{t}(P)$ approximates therefore the degree of coordination in an assemblage as it captures the evolution of both integration and inner distinctiveness of an assemblage. Mechanisms of coordination are therefore foundational to our approach and are further discussed next.

\subsection{Coordination} \label{subsec:coordination}
We understand coordination as \textit{the reciprocal regulation of behavior given in terms of exchanging matter, energy or information among interacting agents, or, between an agent and its environment}. In the latter case, the very distinction of agent -- environment already involves a basic level of coordination. Looking deeper into the nature of interactions among agents at a single stratum $P$ we need to further understand the mechanisms by which populations of agents reduce disparity and incompatibility and progressively individuate towards integrated and coordinated higher-level individuals. These mechanisms were already mentioned briefly in \ref{subsec:phases} as `habit forming'\footnote{The tendency to form habits or repeating patterns of interaction is philosophically profound. It seem to indicate an ontological bias towards coordination over disparity, and more generally, of order over disorder. This goes back to transcendental empiricism being our point of departure. The co-emergence of observer and observed necessarily reflects an intrinsic bias (though temporary and local) towards order over disorder, otherwise neither observers nor observations could possibly emerge. Order, therefore is both self-evident and self-generative and so is the intelligence manifesting in it.}. Such mechanisms, we learn, are local and distributed over the population but need to be capable of achieving effects of global consequences.

 Two major categories of mechanisms can be identified according to the aspect of the system that they affect\footnote{The distinction made here is clear only in the context of a single stratum but is much less apparent considering multiple strata as topological changes in one stratum lead to behavioral changes in the stratum above it.}: \begin{inparaenum}[\itshape a\upshape )]
\item topology modifying mechanisms and
\item behavior modifying mechanisms.
\end{inparaenum}
Topology modifying mechanisms manipulate the relative frequencies of interactions among agents depending on their particular nature. The principle common to such mechanisms is that interactions that contribute to compatibility and coordination will tend to increase in frequency while those contributing to incompatibility will tend to be suppressed. The global topology of the network changes as links between compatible agents will become stronger while links between incompatible agents will become weaker or disappear. 

As a simple example, consider a group of people speaking a number of different languages. When interacting, people will tend to communicate with interlocutors speaking the same language and communication attempts with interlocutors who speak other languages will quickly become infrequent. Also, if there is no choice, people will seek those who speak a language that is similar to theirs and shares some limited vocabulary. Another very well known example is Hebbian learning in networks of neurons where synapses strengthen in correlation to synchronous firing of neurons before and after their synaptic connection \citep{hebb_organization_1968}. The local modifications of topology achieve eventually global effects.           

The significance of topology modifying mechanisms is that interactions taking place over the network cause the modification of the structure of the network. Note also that the topological structure shaped by interactions further affects the future flow of interactions and therefore future global behaviors. By that, topology modifying mechanisms realize a transductive process, as discussed in \ref{subsec:transduction}. 

The second category of mechanisms have to do with behavior modification. Agents can overcome their initial mutual incompatibility and become coordinated by constraining their own or each others' set of possible behaviors depending on their interactions\footnote{For an early fascinating account of the idea of self-organization in the sense described here, see \citep{ashby_principles_1962}.}. In other words, they reciprocally determine or select each others' behaviors and by doing that they bring forth mutual relevance and coordination. 

Mutual modification of behavior requires direct or indirect reflexivity among agents. If agent $A$ affects the behavior of agent $B$, but is not affected, directly or indirectly, by the modifications of behavior it has initiated, there is no real sense in speaking about progressive resolution of disparity. Even in the case that the effects of $A$ on $B$ have reduced the incompatibility between $B$ and another agent $C$, with no feedback to $A$ of this reduction, the influence of $A$ is only contingent and no recurrent pattern can emerge. If however some degree of reflexivity does exist, the exchange can eventually reach a relatively stable and recurrent set of interactions among the participant agents and an operational closure may emerge\footnote{All forms of conditioning including self-conditioning belong to this category as they establish correlations between an agent's input and output signals.}.  

Two observations can be made here. The first is that achieving coordination is primarily a cybernetic\footnote{The cybernetic nature of individuation was already discussed in \ref{subsec:phases} but here it is introduced in the more specific context of our model.} selective process that involves feedback. The second conclusion is somewhat more complex; in order to participate in a coordinated assemblage, agents need to be reciprocally sensitive to the states of each other. We can see now why $Int_{t}(P)$ corresponds approximately to higher degrees of coordination, but we can also see the limitation of $Int_{t}(P)$, since it does not necessarily indicate the bi-directional information exchanges that are necessary to establish recurrent patterns. Supporting these observations is Edelman's discussion and research of re-entrant neural circuits \citep{edelman_neural_1987,edelman_reentry:_2013,tononi_reentry_1992}.

The regulation of interactions whether by constraining the network topology or the actual behavior of the agents can be thought of as a \textit{meta capacity} of agents because they not only affect and are affected by other agents but can also regulate the manner by which they affect and are affected. According to \citet[p. 39]{di_paolo_horizons_2010}, the difference between structural coupling of an agent with its environment (or other agents) and the regulation of this coupling is the definitional property of a cognitive system. But such regulation is not designed. It gradually emerges in the course of interactions that are at least initially contingent. Therefore, we do not see merit in drawing sharp lines between systems that are cognitive and systems that are not when it is evident in many if not all cases that sense-making, the mark of cognition, is a matter of a gradual continuous process of individuation.    

In summary, the underlying processes of sense-making can be understood in cybernetic terms. These are mutually selective processes distributed over populations of interacting agents. They `explore' and spontaneously `discover' novel coordinated interactions among the participating agents. A new sense consolidates however only when such `discovered' coordinated interactions become recurrent (`forming a habit'). The tendency towards the formation of recurrent patterns of interactions is not given \textit{a priori}. It is itself an outcome of individuation as certain coordinated interactions contingently form operational closures or fluid identities that resist change to a greater or lesser degree. If there was absolutely no such tendency, there could be no coordination, no individual objects or persistent relations between objects, just disorder. 

\subsection{Perspective and value} \label{subsec:perspective}
The concept of value occupies a primary place in the discourse about the nature of intelligence. In \ref{subsec:current_definition}, the ability of an agent to achieve goals is mediated by maximizing rewards. The combination of a goal and environment create for the intelligent agent a perspective by which all situations whether internal or external, and all agent -- environment interactions, gain significance in terms of how they reflect on the achievement of the goal. Values can be generally described as the quantitative measures of significance and the dynamics of values guide the actions of the agent. In his analysis of intelligent agents \citet{legg_machine_2008} writes:
\begin{quotation}
[...] We define an additional communication channel with the simplest possible semantics: a signal that indicates how good the agent's current situation is. We will call this signal the reward. The agent simply has to maximize the amount of reward it receives, which is a function of the goal. In a complex setting the agent might be rewarded for winning a game or solving a puzzle. If the agent is to succeed in its environment, that is, receive a lot of reward, it must learn about the structure of the environment and in particular
what it needs to do in order to get reward. \citep[p 72]{legg_machine_2008}
\end{quotation}
Traditionally, intelligence is measured in terms of finding ways to maximize the reward (value) for various environments and goals. Of course, the value function itself may be subject to changes in time and additionally, strategies that consider short-term or long-term maximum rewards might be profoundly different. Still, as the commonly accepted concept of intelligence is understood, the manipulation of measurable value by the agent is what intelligence is all about and therefore value must be a given (see also \ref{subsec: criticism}). 

The enactive theory of cognition follows a similar approach but with two important differences:
\begin{inparaenum}[\itshape a\upshape )]
\item specific values are not \textit{a priori} given but are self-generated by an operational closure and characterize an autonomous identity: ``Sense making: Already implied in the notion of interactive autonomy is the realization that organisms cast a web of  significance on their world.''\citep[p. 39]{di_paolo_horizons_2010}(see also \ref{subsec:phases}), and
\item the preservation of identity is the prime value of autonomous systems: ``For enactivism, value is simply an aspect of all  sense-making, as sense-making is, at its root, the evaluation of the consequences of interaction for the conservation of an identity.''\citep[p. 45]{di_paolo_horizons_2010}. 
\end{inparaenum}
Indeed according to enactivism, specific value functions are not given, but there is a primal value which is the conservation of identity. Di Paolo et al. later define value as ``[ ] the extent to which a situation affects the viability of a self-sustaining and precarious network of processes that generates an identity.'' \citep[p.48]{di_paolo_horizons_2010}, which makes it even clearer that there is an \textit{a priori} value in place. In both approaches, value guides behavior but is also a limit. Once achieved or maximized, the potential of the agent for further exploration is exhausted. 

Though we accept that values are intrinsic to sense-making, we do not agree that values must precede any intelligent activity or sense-making in order to guide them; nor that they are necessarily preceded by the establishment of an autonomous identity. Rather, values are products of an ongoing individuation. Emerging values in the process of individuation carry their own problematic as they are initially non-coherent or even conflicting. A good example of the individuation of value is the negotiation over the price of a certain good in a marketplace. If the market is big enough, and the good is offered by a few vendors, the price of the same good can be negotiated in many places by different agents and reach significantly different values. However, information exchanged among buyers and sellers over the whole market will eventually minimize or eliminate the variation in the price\footnote{Such processes of individuation can become extremely complex. This example also demonstrates that considering a single price for a good is often a gross oversimplification. Prices of goods undergo an individuation process that is never exhausted especially if demand and supply are distributed and fluctuating.}. When in the course of individuation, values become relatively invariant, they become the characteristics of stable individuals. The effective regulation of such values by individuals can then be understood as the preservation of identity. Values can designate a certain relation or set of relations (e.g. body temperature relative to the environment, the skin color of a chameleon etc) between an agent and its environment. When such value becomes regulated and therefore relatively stable, it guides general categories of behaviors such as adaptation (i.e. the modification of internal structure in response to perturbations), or niche construction (i.e. the modification of the structure of the environment in response to perturbations). The development of behaviors that belong to these categories are well accounted for by the conventional conception of intelligence. 

Regulated values, by definition, resist change. Therefore it is easy to understand why they are associated with identity. Identity is nothing more than a set of variables being kept within a certain range of values. Identity and values therefore co-define each other. Values in the course of individuation, in contrast, cannot be said to characterize an identity. In fact, they cannot be conventionally identified as values at all. In the preindividual state there are no values, only \textit{proto-values}. 

In the multi-strata model of individuation we describe in \ref{subsec:descriptive}, every stratum individuates its own set of values that also reflect the different identities of agents that emerge in that stratum. The individuation of agents in any stratum $S$, however, does not depend only on horizontal interactions within that stratum. The relations of every stratum with its substratum $S_{sub}$ and superstratum $S_{sup}$ is mediated by values that emerge in these neighboring strata. The substratum $S_{sub}$ provides the component elements that constitute the assemblages in $S$. Inasmuch as these elements are more or less stable individuals, they have characteristic values that resist changes and perturbations that may be caused by interactions in $S$. In other words, the values that emerged at the substratum $S_{sub}$ are selective (i.e. constraining) in regard to the interactions possible in $S$. In a similar manner, individuations that take place on the superstratum $S_{sup}$ will tend to regulate the individuations on $S$ by preferring certain interactions over others. For example, if an agent produced in $S$ is frequently involved in assemblages emerging in $S_{sup}$, this will have a biasing effect on the distribution of agents within the population of agents in $S$. Changing the distribution of agents in the population exerts certain constraints as well as allowing certain degrees of freedom on the interactions taking place in $S$. 

In summary, individuation at every stratum is subject to both bottom-up and top-down influences that are mediated by the values in neighboring strata. Individuation at multiple simultaneous levels involves both evolutionary (bottom-up) and developmental (top-down)
organization. For example: a human organism in a social context is exposed to systems of individuating pressures that in turn affect biological parameters (e.g. stress) that affect the individuation of specific organs that in turn affect the individuation of cell populations and individual cells. A cell may produce a mutation, undergo destabilization of its genetic operations as a result, and turn into a cancerous cell that is as stable as an healthy cell. This may disrupt a tissue or a whole organ and affect the performance of the affected human in her social context (e.g. disability and need for medical care). The division into strata reflected in our model is not an artificial construction though. It derives from the fact that complex individuation processes spontaneously produce an hierarchy of individuated entities because low-level simple assemblages are more probable (and therefore faster) to integrate into coordinated wholes than complex assemblages. This results in the emergence of a stratified process of individuation. See \citep{simon_architecture_1962} for further discussion of this effect.   

\section{A non-concluding conclusion} \label{sec:discussion}
Cognitive science and artificial intelligence research have made very impressive advances, in understanding and practically implementing systems with a wide range of intelligent capacities. Yet, most of the current theoretical thinking about intelligence and cognition is still limited to a problem solving dogma, as argued in \ref{subsec:current_definition}-\ref{subsec: criticism}. In this paper we go beyond the identifiable cognitive competences that can be readily associated with specific problems or problem domains. We lay down philosophical and theoretical foundations to how intelligent systems such as brains, whole organisms, social entities and other organizations develop and scale. We shift the focus of investigation from intelligent agents as individual products to the intelligence intrinsic to their process of production i.e. their individuation -- what we call open-ended intelligence. We propose that such an approach provides important insights as to what  differentiates intelligence that is open-ended and truly general from other goal oriented and therefore limited types of intelligence. By that, we offer a significant extension to the conceptualization and understanding of intelligence.

The principle distilled from this investigation is that \textit{Open-ended Intelligence is a process where a distributed population of interacting heterogeneous agents achieves progressively higher levels of coordination. In coordination here we mean the local resolution of disparities by means of reciprocal determination that brings forth new individuals in the form of integrated groups of agents (assemblages) that exchange meaningful information and spontaneously differentiate (dynamically and structurally) from their surrounding milieu.} This kind of intelligence is truly general in the sense that it is not directed or limited by an \textit{a priori} given goal or challenge. Moreover, it is intrinsically and indefinitely scalable, at least from a theoretical point of view. We see open-ended intelligence manifesting all around us and and at many scales; primarily in the evolution of life, in the phylogenetic and ontogenetic organization of brains, in life-long cognitive development and sense-making and in the self-organization of complex systems from slime molds, fungi, and bee hives to human sociotechnological entities. 

Interestingly, open-ended generative intelligence is reflexively involved in the very process of describing it here in the individuation of concepts, models and perspectives explored above. And these, we learn, are always a work in progress. We conclude this paper therefore by highlighting problems and disparities in the form of a few challenging open questions that stimulate further research and may drive further individuation.
\begin{description}
	\item[Measuring~open-ended~intelligence] -- The goal-oriented approach to General Intelligence is particularly successful in providing a simple and reliable measure of fitness or success that can be directly associated with the level of intelligence an agent presents. In our case however, measurement is much less obvious. In order to have a better grasp of the dynamics of intelligent individuating processes, more rigorous measures of individuation need to be developed. Because of the unique nature of individuation as a determining process it is not entirely clear whether or not it can be generally quantified. Our point of departure for measurement is the concept of information integration that was developed by Tononi \citeyearpar{tononi_information_2004,tononi_consciousness_2008} in a neuroscientific context as a possible explanation of consciousness. We use this concept in a somewhat different and more general way to quantify individuation. Measures based on information integration derive only from the probabilistic properties of the exchanged signals (appendices A and B sketches preliminary steps in that direction). While this might be sufficient for low level agents such as neurons, or similarly simple agents, they do not capture the full significance of affect between more complex interacting elements (e.g. human decision makers) in the general case. This must necessarily involve a notion of the meaning embedded in the exchanged signals (i.e. what difference do they make for the agents). In other words, information integration is not sufficient to express the manner by which elements within an assemblage actually affect and are affected by each other. They merely reflect that such affective relations are taking place and to what extent. In order to quantify open-ended intelligence, a measure must be developed to reflect the degree of coordination achieved within a population of agents at each stage of individuation. Additionally, a measure needs to be developed to estimate relative stability and resilience of already formed individuals within a population. Such measure(s) will allow us to better understand and monitor the dynamics of individuation, turning points, disruptive elements and more. 
	\item [Value~systems~and~stratified~individuation] -- Of special interest is to investigate the individuation of values. Values represent consolidated goals and are therefore highly significant in understanding the evolution of intelligent competences and sense-making. Values are signals that guide distinction mechanisms thus enabling adaptation and learning. In our understanding, values also mediate between different strata of individuation. The individuation of values seems to be an important key to further understand the individuation of intelligent systems across strata.	
	\item[Towards~a~generative~model] -- One of the more difficult and interesting challenges is to implement a simulation model of open-ended intelligence based on the concepts explored in this paper. Such an implementation will serve both theoretical and practical ends. It will help to better understand individuation and the transduction mechanism and it will help to understand or even discover general coordination mechanisms. It will help to appreciate the potential and limitations of scalability, and whether truly open-ended systems are practically possible and under what conditions. Importantly, it may also become a platform for specific applications. 
	\item[Understanding~coordination] -- We see the individuation of coordination as the manifestation of open-ended intelligence. One of the focuses of future research would be to investigate individuation processes in the light of the kinds of coordination they bring forth. For example: synchronization is a very basic type of coordination having to do with the timing of activities and recurrent patterns of interaction. The phenomena of resonance is instrumental to understanding how agents that are initially not synchronized (i.e. disparate in the temporal sense) can gradually synchronize their interactions. Another important topic is to investigate the relations between the individuation of coordination within stratum and across strata.		
	\item[Potential~for~application] -- Observing individual systems, we are often able to see in retrospect that the system evolved to address a specific problem (e.g. eyes, flowers, wings, courts, money, transportation systems, the Internet etc.). But it is very difficult, if at all possible, to foresee what final purpose or goal a system might fulfill while it is individuating, when it is not a system as yet and the interactions among its prospective future components carry only marginally meaningful information. Open-ended intelligence therefore seems to be inherently unpredictable as to its final products and as a result difficult to be harnessed towards a useful purpose. It will be interesting to investigate the possible practical applications of individuating processes and whether they can be guided \citep{prokopenko_guided_2009}. Of interest is also the hybridization of goal oriented and individuating approaches to achieve highly fluid intelligent systems.  
\end{description}
	
\printbibliography

\pagebreak

\section*{Appendix A} \label{app:appendixA}
\subsection*{Information integration as a measure of boundary formation in a population of interacting agents}
Given a population $P$ of $p_{i}$ interconnected agents, where $i \in [1,..,N]$, we wish to quantify how much they affect and are affected by each other. In information theoretic terminology, each agent $p_{i}$ can either change its state independently of all other agents in $P$, or its state may depend on the states of other agents in $P$, or even be entirely determined by the states of other agents. The mutual information between two agents $p_{i},p_{j}$ is given by the formula: 
\begin{eqnarray}
	MI(p_{i},p_{j})=H(p_{i})-H(p_{i}/p_{j})=H(p_{j})-H(p_{j}/p_{i})\\
	=H(p_{i})+H(p_{j})-H(p_{i},p_{j})
\end{eqnarray}

Where $H(x)$ is the entropy involved in the state of agent x. If $p_{i}$ and $p_{j}$ are independent, $H(p_{i},p_{j})=H(p_{i})+H(p_{j})$ and then $MI(p_{i},p_{j})$ would be 0. The mutual information would be maximum in the case that the state of one agent is fully determined by the other. In this case the mutual information will be equal to $min(H(p_{i}),H(p_{j}))$.

For a set of agents $p_{i}$ in $P$ the integration of the whole set would be given by the sum of the entropies of the independent agents $p_{i}$ minus the entropy of the joint set $P$:
\begin{equation} \label{eq:integration}
	I(P)=\sum_{i=1}^{k}H(p_{i})-H(P)	
\end{equation}

In order to compare the degree of integration within a subset of agents to the integration between the said subset and the rest of the population, we divide the population of agents $P$ into two subgroups of differing sizes: $X_{i}^{k}$ and its complement $P-X_{i}^{k}$, where k is the number of agents in the subset $X$. The mutual information between $X_{i}^{k}$ and its complement is:
\begin{equation} \label{eq:MI}
	MI(X_{i}^{k},P-X_{i}^{k})=H(X_{i}^{k})+H(P-X_{i}^{k})-H(P)
\end{equation}

Formula \ref{eq:MI} measures the statistical dependence between a chosen subset $i$ of $k$ agents and the rest of the population. The \textit{Cluster Index $CI$} of the subset $X_{i}^{k}$ will therefore be given by:
\begin{equation} \label{eq:cluster}
	CI(X_{i}^{k})=I(X_{i}^{k})/MI(X_{i}^{k},P-X_{i}^{k})
\end{equation}
$CI$ measures the degree of distinctiveness of a subset of agents in $P$ compared to the whole population in terms of information exchange\footnote{Note that these are only simplified formulas that do not take into account the different sizes of subsets.}. For $CI\le1$ there is no significant distinctiveness while a subset with $CI\gg1$ indicates a distinct integrated cluster. A threshold on $CI$ can therefore be used to formally describe more or less integrated assemblages.  

\section*{Appendix B} \label{app:appendixB}
\subsection*{The operational complexity of a system of interacting agents in a population}
A simplified general measure of operational complexity can be given in terms of the average mutual information of subsets of $P$. Let $P$ be a population of size $M$. Assume that $P$ is isolated so its inner states are self produced. We divide $P$ into two complementary subsets $X_{j}^{k}$ and $P-X_{j}^{k}$ of respective sizes $k$ and $M-k$. The index $j$, enumerates all possible subsets of size $k$ out of $X$. The operational complexity $OC(P)$ of population $P$ can be given by:
\begin{equation} \label{eq:operational_complexity}
	OC(P)=\sum_{k=1}^{M/2}<MI(X_{j}^{k},P-X_{j}^{k})>
\end{equation} 
where the mutual information is averaged on all subsets of size $k$. Subsets of very small size will contribute very little to $OC(P)$, while subsets of sizes in the vicinity of $M/2$ will contribute the most complexity. Remarkably, $OC(P)$ measure of complexity is based only on the extent to which subsets of the population affect each other and the statistical properties of the signals that agents within the population exchange. $OC(P)$  therefore does not rely on an arbitrary measure of complexity imposed from outside the cluster. 

\end{document}